\documentclass[conference]{IEEEtran}
\IEEEoverridecommandlockouts    
\usepackage{cite}
\usepackage{amsmath,amssymb,amsfonts}
\usepackage{algorithmic}
\usepackage{graphicx}
\usepackage{textcomp}
\usepackage{xcolor}
\usepackage{tikz}
\usetikzlibrary{shapes,positioning,calc,arrows,chains}
\usepackage{adjustbox}
\def\BibTeX{{\rm B\kern-.05em{\sc i\kern-.025em b}\kern-.08em
    T\kern-.1667em\lower.7ex\hbox{E}\kern-.125emX}}
\begin{document}

\title{Impact of Tokenization on \\ LLaMa Russian Adaptation%\\
}

\author{\IEEEauthorblockN{1\textsuperscript{st} Mikhail Tikhomirov}
\IEEEauthorblockA{
\textit{Lomonosov Moscow State University}\\
\textit{Moscow, Russia} \\
tikhomirov.mm@gmail.com}
\and
\IEEEauthorblockN{2\textsuperscript{nd} Daniil Chernyshev}
\IEEEauthorblockA{\textit{Lomonosov Moscow State University}\\
\textit{Moscow, Russia} \\
chdanorbis@yandex.ru}
}

\maketitle
\begin{tikzpicture}[remember picture, overlay]
\node at ($(current page.south) + (0,0.65in)$) {
\begin{minipage}{\textwidth} \footnotesize
 \copyright~2023 IEEE. Personal use of this material is permitted. Permission
 from IEEE must be obtained for all other uses, in any current or future media,
 including reprinting/republishing this material for advertising or promotional
 purposes, creating new collective works, for resale or redistribution to
 servers or lists, or reuse of any copyrighted component of this work in other
 works.
\end{minipage}
};
\end{tikzpicture}
\begin{abstract}
Latest instruction-tuned large language models (LLM) show great results on various tasks, however, they often face performance degradation for non-English input. There is evidence that the reason lies in inefficient tokenization caused by low language representation in pre-training data which hinders the comprehension of non-English instructions, limiting the potential of target language instruction-tuning. In this work we investigate the possibility of addressing the issue with vocabulary substitution in the context of LLaMa Russian language adaptation. We explore three variants of vocabulary adaptation and test their performance on Saiga instruction-tuning and fine-tuning on Russian Super Glue benchmark. The results of automatic evaluation show that vocabulary substitution not only improves the model's quality in Russian but also accelerates fine-tuning (35\%) and inference (up to 60\%) while reducing memory consumption. Additional human evaluation of the instruction-tuned models demonstrates that models with Russian-adapted vocabulary generate answers with higher user preference than the original Saiga-LLaMa model.
\end{abstract}

\begin{IEEEkeywords}
large language models, tokenization, language adaptation, llama
\end{IEEEkeywords}

\section{Introduction}
With the introduction of powerful pre-trained language models such as BERT~\cite{Devlin2019}, neural models became the basis of Natural Language Processing (NLP) solutions. Thanks to accumulated global contextual knowledge the pre-trained language models exhibit higher task adaptability, requiring less training data and computational resources for fine-tuning while achieving better results. However such models are harder to replicate as the computational costs of the pre-training procedure are several times higher than training the model from scratch on downstream task~\cite{Brown2020}, which prompted the researchers to pursuit the universal model with the best out-of-box performance.  

The initial works focused on task-specific pre-training strategies~\cite{Yoon2019,Zhang2020,Tikhomirov2020}, however, the exploration of knowledge accumulation boundaries led to the discovery of a more flexible alternative, in-context learning~\cite{Brown2020}. In contrast to conventional fine-tuning approaches, in-context learning adapts to the task by conditioning on task description (instruction) and a set of analogous examples that are provided within the user input which tunes the network attention mechanism to the task relevant knowledge. This capability was shown to appear only in larger language models (over 6B parameters) that have a sufficient network capacity to memorize the context of various domains and tasks. Latest iteration of in-context learning leverages explicit instruction tuning~\cite{Ouyang2022} to eliminate the need for analogous examples reinforcing the role of Large Language Models (LLM) as universal NLP task solvers. 

Current state-of-the-art instruction-tuned LLMs such as ChatGPT show outstanding zero-shot performance on various domains, yet their performance significantly degrades when applied to non-English domains. The main reason is the dominance of English data in available pre-training datasets which leads to English bias in both knowledge and morphology. Consequently, developing a language-specific counterpart is much more challenging due to lack of domain coverage which is required for achieving model universality. A more affordable alternative is fine-tuning multilingual LLMs on target-language instruction datasets~\cite{Cui2023}. Most implementations\footnote{https://github.com/avocardio/Zicklein}\footnote{https://github.com/22-hours/cabrita}\footnote{https://github.com/IlyaGusev/rulm} use LLaMa model~\cite{Touvron2023} as the backbone for instruction-based language adaptation, however, being an English-focused model it suffers from low tokenization efficiency for other languages. Beside performance overhead related to excessive number of tokens, inefficient tokenization also results in violation of morphological characteristics, hindering the language-adaptation process~\cite{Nayak2020,Hofmann2021}. 

We argue that tokenization optimization is the necessary step for efficient LLM language adaptation. In this work, we explore different embedding optimization variants in context of LLaMa model and compare their performance on fine-tuning and instruction-tuning settings. Following generalized vocabulary substitution method \cite{Kuratov2019, Rust2021} we considered three vocabulary options: Russian BPE, Russian Unigram and the original. Additionally, we analyzed two different algorithms for embedding and LM head initialization. The quality evaluation was done on Russian Super Glue benchmark and side-by-side tests. Our key contributions are:
\begin{itemize}
    \item We show that Unigram tokenization has higher morphological accuracy than the BPE algorithms, utilized in state-of-the-art models;
    \item Our benchmark demonstrates that Russian large language models highly benefit from morphologically accurate word tokenization, achieving a considerable quality boost with Unigram vocabulary at all evaluation settings;
    \item Additional Human evaluation performed by 15 annotators reveals that vocabulary substitution significantly improves the instruction-tuning efficiency, with better relevance of generated answers;
    \item Performance tests indicate that vocabulary substitution substantially boosts the efficiency of resource utilization, accelerating model inference by up to 60\% while also reducing the memory consumption.
\end{itemize}
\section{Related Work}
Embedding adaptation has been popular technique since the introduction of pre-trained language models. Lakew~\cite{Lakew2018} et al. studied the problem in the scope of cross-lingual neural machine translation transfer learning and showed that adapting input vocabulary to unseen languages could be more efficient that training a full model from scratch. Similarly, Kuratov et al.~\cite{Kuratov2019} showed that vocabulary substitution of multilingual BERT and pre-training the updated embeddings significantly improves the model performance on Russian language. A more detailed study of this language optimization approach was done by Rust et al~\cite{Rust2021}, which concluded that the performance improvements generalize to all languages. Later Vries et al.~\cite{Vries2021} adopted the vocabulary substitution approach for GPT-2 cross-lingual adaptation and achieved comparable results to Italian-specialized counterpart. Zeng et al.~\cite{Zeng2022} improved the vocabulary substitution approach for monolingual models with bilingual lexicon embedding initialization, which allowed to skip the embedding pre-training stage for encoder-only pre-trained models.

\section{Methodology}
Following previous works\cite{Kuratov2019, Rust2021} we use generalized vocabulary substitution training strategy to adapt LLaMa model to Russian language. The approach can be summarized in the following steps:
\begin{enumerate}
    \item Build a new tokenization vocabulary on target language corpus,
    \item Rebuild the embedding and LM head layers to account the new vocabulary size,
    \item Initialize the updated layers using the old embedding weights in accordance to the overlap between old and new tokenizations,
    \item Pre-train the updated layers on language modeling task using the target language corpus.
\end{enumerate}
The implementation detatils of each step will be discussed further.

\subsection{Step 1: Tokenization}
We considered BPE and Unigram tokenization algorithms. Both algorithms use subword vocabularies (char n-grams), however, they differ in terms of vocabulary construction strategies. BPE uses bottom-up strategy, merging the subword tokens according to their corpus frequency. In contrast, Unigram is top-down approach that constructs the tokens by pruning them to maximize the corpus likelihood. For that reason Unigram is more prone to retain elementary language semantic units like roots or stems~\cite{Bostrom2020}. We hypothesize that this trait might be beneficial for inflected languages such as Russian.

\subsection{Step 2-3: Embedding initialization}
To initialize replaced matrices, new tokens were processed based on the old tokenization, and then the corresponding rows of the matrix were initialized by averaging the vectors of resulting tokens in the original embedding matrix. In particular, in order to implement this procedure correctly, we need to carefully handle tokens without a leading space, since the tokenizer by default treats the first word as having a space before it. 
\begin{equation}
\begin{array}{c}
    v_{new}(t_i^{n}) = \frac{1}{K}\sum_{j=1}^{K}v_{raw}(t_j^{r}); \\ \ \\
    tokenize_{raw}(t_i^{n}) = [t_1^{r}, ..., t_K^{r}],
\end{array}
\end{equation}
where $tokenize_{raw}$ is original tokenization function, $t_i^{n}$ - token in new tokenization, $t_1^{r}$ - token on original tokenization, $v_{raw}(...)$ - vector of token in original model.

To initialize LM head, we investigated 2 options: 1) initialization of LM head with a copy of the embedding layer, and 2) initialization by analogy with embeddings, but using the old LM head (we will call such a case with the suffix \textit{hm}).

\subsection{Step 4: Continued target-language pre-training}
We assumed that for language adaptation of multilingual LLM's it is enough to train only the embedding layers, since the model had to develop universal language semantics during it's pre-training. For this reason, we froze all layers except the embedding and the LM head and trained on a Russian-language text corpus on the causual language modeling task (the pre-training task of the original LLaMa model) with cross-entropy loss. This decision also makes training procedure more affordable, as the number of training parameters are no more than a few percent of the total model size, thus alleviating memory requirements.

\section{Experimental setup}
\subsection{Training data}
For our adaptation experiments we collected documents from the following domains:
\begin{itemize}
    \item Russian Wikipedia
    \item Pikabu
    \item News
    \item Habrahabr
    \item Stackoverflow
    \item Books
\end{itemize}

The documents were deduplicated using Locality Sensitive Hashing Minhash algorithm. To better model Russian language vocabulary distribution and to avoid grammatically incorrect examples the metadata, links and comment sections of web-pages were dropped. We used UTF-8 normalization to remove non-standard symbols like emoji or logograms (e.g. Chinese characters), thus restricting the vocabulary to Cyrillic and Latin languages. Text structure was standardized to singular space characters (i.e. no multi-spaces or multi-newlines).  After all filtering procedures we obtained a dataset\footnote{Link to dataset is hidden for the purpose of blind review} of 9 million documents or 3.5 billion words ($\sim$33GB). 

\subsection{Tokenization}

To train the tokenizers we used sentencepiece package\footnote{https://github.com/google/sentencepiece} on a subset of 5M documents from our language adaptation dataset. The tokenization characteristics were evaluated on RuMorphs-Words\footnote{https://cmc-msu-ai.github.io/NLPDatasets/} dataset, which contains morphological labels for more than 1.7M Russian words. To measure morphological accuracy we examined root integrity which is defined as the maximal length of longest common char subsequence among all word tokens normalized by root length. 

\begin{figure}
    \centering
    \includegraphics[width=0.5\textwidth]{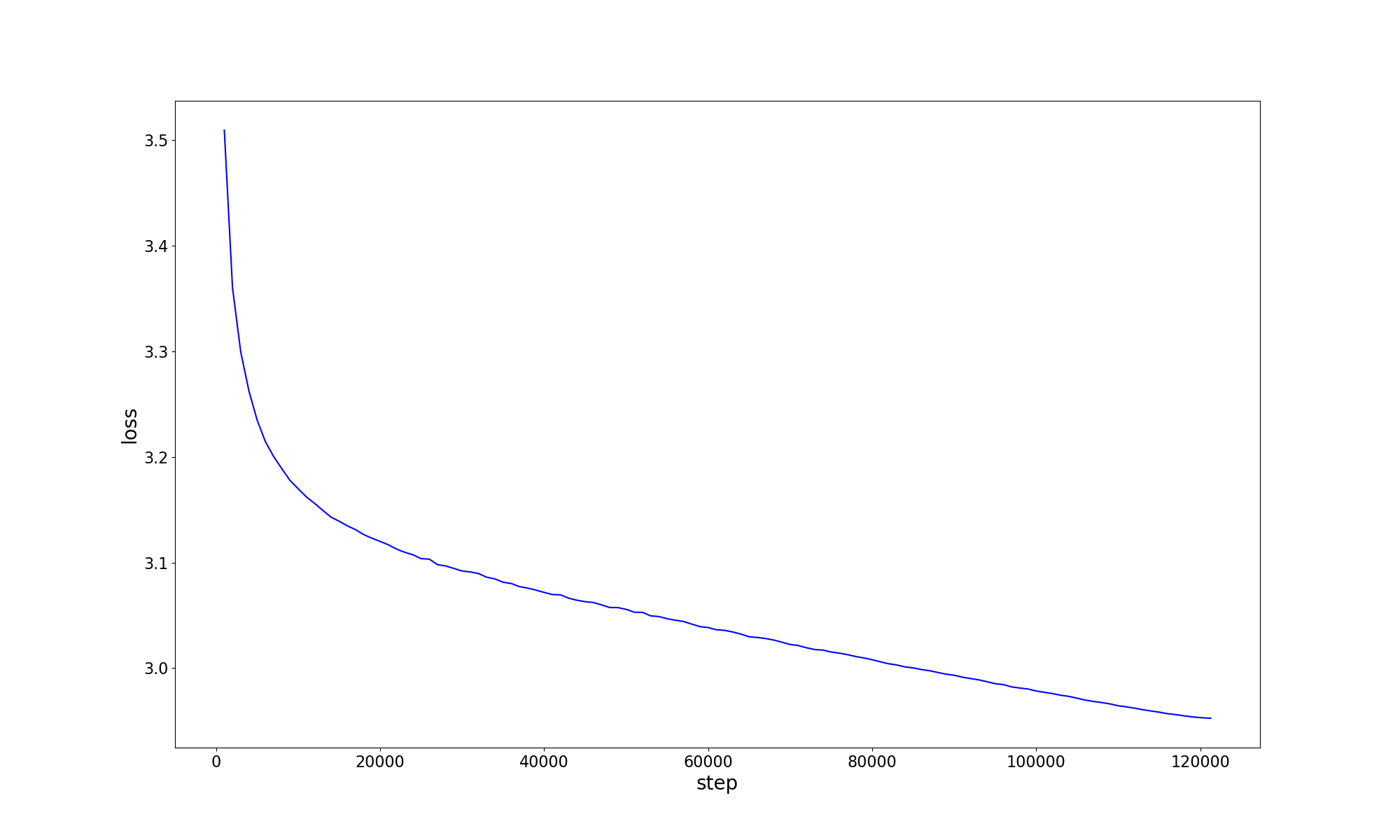}
    \caption{Validation loss when adapting on Russian language}
    \label{fig:eval_loss_adaptation}
\end{figure}

\subsection{Training settings}
Training of new layers on the Russian language modeling dataset took place on 12 Nvidia V100. Since LLaMa-7b isn't natively supported by Volta architecture we had to use DeepSpeed Zero-2 to accelerate the training. The text data was chunked with Block size = 128 without word boundary alignment. We used Adam optimizer with lr = 3e-04, linear scheduler with warmup of 2000 steps and total batch size = 240. To prevent overfitting the models were trained with early stopping and evaluation on validation set every 0.1 epochs.

Our preliminary experiments indicated that 1 epoch was sufficient for achieving the original level of language coherence. Therefore, to save the costs and to ensure the purity of the experiment all variants of language adaptation were trained for that duration. However, it must be noted that 1 epoch is suboptimal (Figure \ref{fig:eval_loss_adaptation}) and the models could be improved by further pre-training.

\subsection{Evaluation protocol}
As the baseline we compared against the original LLaMa-7b model and it's continued pre-training version (designated with the suffix \textit{raw}) that was embedding-tuned without vocabulary substitution on our Russian pre-training corpus. The latter is necessary to measure the net quality increase from the improved tokenization and to disprove the hypothesis that the continued pre-training on it's own could be optimal for language adaptation.

To evaluate language modeling performance we used  Russian Super Glue (RSG) benchmark~\cite{shavrina2020} which consists of 9 natural language understanding tasks. To maintain consistency with the existing LLaMa submissions we used evaluation protocol from Saiga model github repository\footnote{https://github.com/IlyaGusev/rulm}. Following the protocol all models were evaluated in two settings: downstream task fine-tuning and instruction-tuned zero-shot. 

In the case of zero-shot, the model was first fine-tuned on the Saiga instruction datasets\footnote{https://huggingface.co/collections/IlyaGusev/saiga-datasets-6505d5c30c87331947799d45} using the LoRA adapter-tuning approach \cite{hu2021}, and then evaluated on RSG tasks through instruction-guided generation.

\begin{figure}
    \centering
    \includegraphics[width=0.5\textwidth]{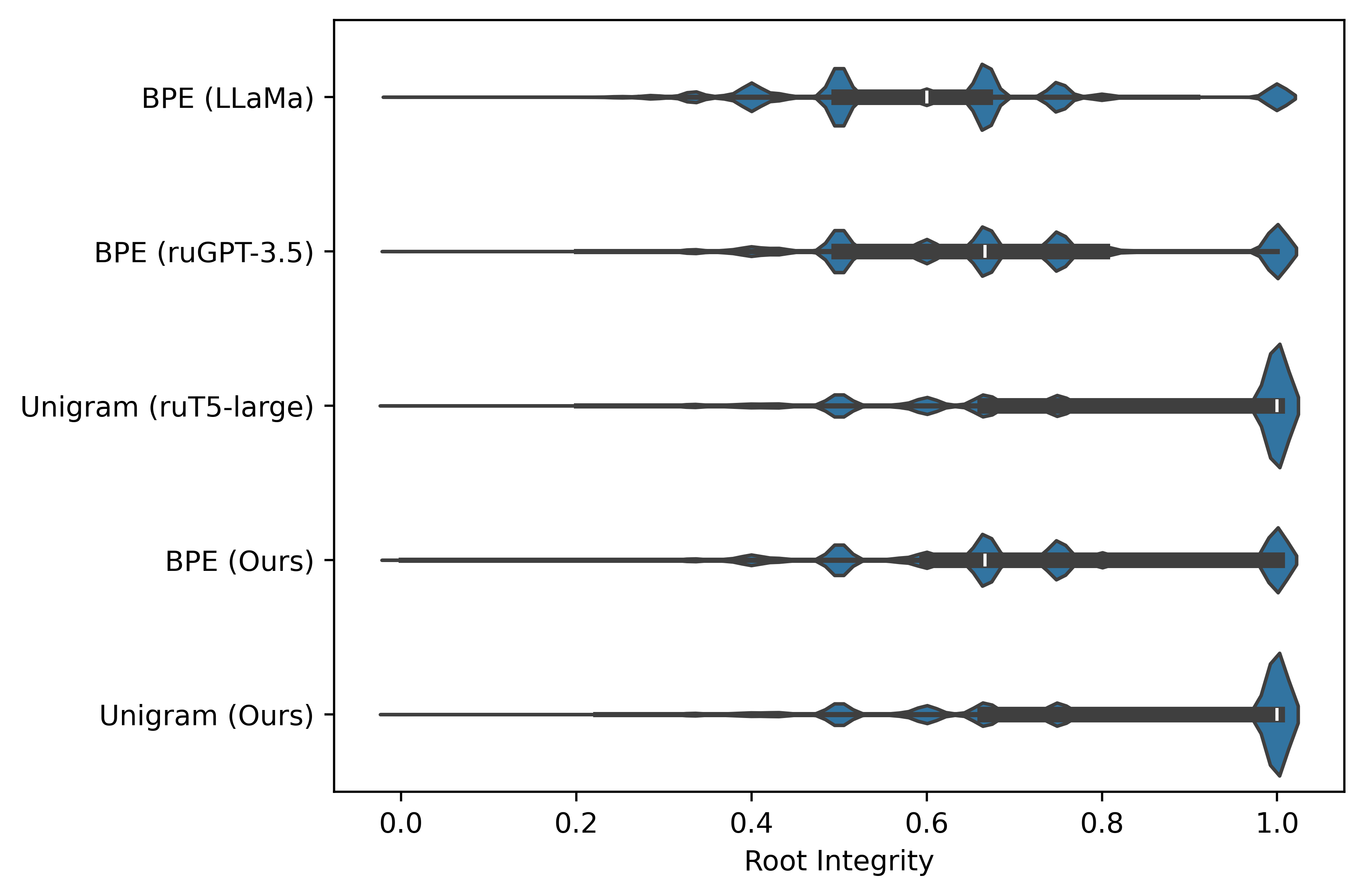}
    \caption{Average root word preservation rate for different tokenizations}
    \label{fig:tok_roots}
\end{figure}

\begin{figure}
    \centering
    \includegraphics[width=0.5\textwidth]{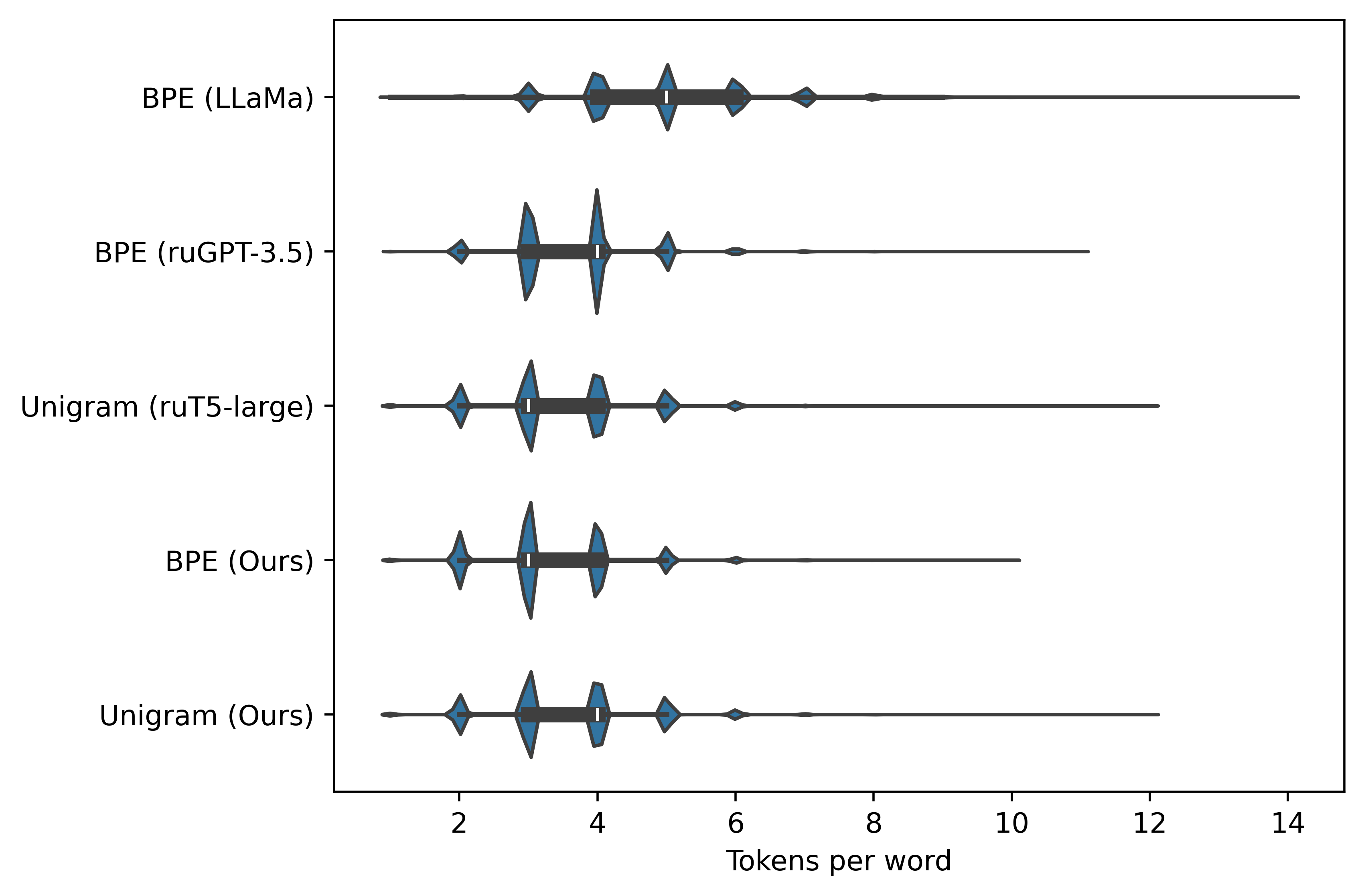}
    \caption{Average word length in tokens for different tokenizations}
    \label{fig:tok_len}
\end{figure}

\section{Results and analysis}
\subsection{Tokenization quality}

Beside the original LLaMa tokenizer we compared to tokenizers of state-of-the-art Russian Language models\footnote{https://huggingface.co/ai-forever/ruGPT-3.5-13B}\footnote{https://huggingface.co/ai-forever/ruT5-large}. According to Figure \ref{fig:tok_roots} conventional BPE algrorithm have lower root preservation rate than Unigram regardless of selected vocabulary. However, the token-wise word length distribution (see Figure \ref{fig:tok_len}) is more biased to lower token count for BPE tokenizer variant, which suggests superior resource efficiency in terms of memory and processing time. At the same time the difference in median token count values between BPE and Unigram isn't enough to justify the significantly lower morphological qualities of the former. Therefore we proceeded with both tokenization variants in our language adaptation experiments. 
\subsection{Russian Super Glue} 
To evaluate the quality of the adapted models, we used the Russian Super Glue benchmark.  The models were assessed in two settings: 1) zero-shot for instruction-tuned versions and 2) fine-tuning on training splits of Russian Super Glue datasets. All fine-tuning (and instruction-tuning) took place using the LoRA approach \cite{hu2021} on Q, V, K, O matrices of attention. In both cases, the answers from the model is obtained by generating text and then automatically searching for key phrases (such as Yes, No) or using the generated text as the entire answer, in the case of question-answering tasks. 

In the case of fine-tuning on Russian Super Glue, all training sets of 9 datasets were combined into a single dataset in a dialogue format.  This approach may not be optimal in terms of achieving the best quality on Russian Super Glue, but it has been used to replicate past results of LLaMa models. Additionally, it was more important for us not to achieve maximum quality, but to examine the difference when using different weights: original and adapted.

The results of both approaches are presented in Tables \ref{tab:1} and \ref{tab:2}. Models based on unigram tokenization perform best on average. BPE tokenization also shows an increase in quality in the zero shot setting, but is inferior to the unigram model. Calibrating the original model for Russian-language data (lama7b\_rulm\_raw and saiga7b\_rulm\_raw) also improves the quality of the model, but is inferior to options with a replacing the tokenization.

\begin{table*}[t]
\centering
\caption{LoRA fine-tune results on Russian Super Glue}

\label{tab:1}       % Give a unique label
%\resizebox{\textwidth}{!}{ 
%\begin{adjustbox}{max width=0.7\textwidth,max totalheight=\textheight,keepaspectratio}
%\resizebox{0.5\textwidth}{!}{ 
\begin{tabular}{l|c|c|c|c|c|c|c|c|c|c}
 & LiDiRus & RCB & PARus & MuSeRC & TERRa & RUSSE & RWSD & DaNetQA & RuCoS & mean \\
\hline
llama7b & 0,361 & 0,462 & 0,672 & 0,799 & 0,860 & 0,624 & 0,682 & 0,866 & \textbf{0,802} & 0,681 \\
llama7b\_rulm\_raw & 0,392 & 0,494 & 0,688 & 0,805 & 0,859 & 0,631 & 0,669 & \textbf{0,871} & 0,791 & 0,689 \\
llama7b\_rulm\_bpe & 0,365 & 0,509 & 0,684 & 0,782 & 0,844 & 0,626 & \textbf{0,747} & 0,824 & 0,737 & 0,680 \\
llama7b\_rulm\_unigram & \textbf{0,412} & \textbf{0,561} & 0,732 & 0,800 & \textbf{0,875} & \textbf{0,660} & 0,675 & 0,865 & 0,756 & \textbf{0,704} \\
llama7b\_rulm\_unigram\_hm & 0,387 & 0,546 & \textbf{0,750} & \textbf{0,815} & 0,866 & \textbf{0,660} & 0,740 & 0,812 & 0,758 & \textbf{0,704} \\

\hline
\end{tabular}

%}
%\end{adjustbox}
%}
\end{table*}

\begin{table*}[t]
\centering
\caption{Zero Shot results of Saiga instruction-tune on Russian Super Glue}

\label{tab:2}       % Give a unique label
%\resizebox{\textwidth}{!}{ 
%\begin{adjustbox}{max width=0.7\textwidth,max totalheight=\textheight,keepaspectratio}
%\resizebox{0.5\textwidth}{!}{ 
\begin{tabular}{l|c|c|c|c|c|c|c|c|c|c}
 & LiDiRus & RCB & PARus & MuSeRC & TERRa & RUSSE & RWSD & DaNetQA & RuCoS & mean \\
\hline
saiga7b & 0,084 & 0,412 & 0,528 & 0,311 & 0,514 & 0,484 & \textbf{0,675} & 0,676 & 0,319 & 0,445\\
saiga7b\_rulm\_raw & 0,025 & 0,373 & \textbf{0,610} & 0,310 & 0,523 & \textbf{0,587} & 0,584 & 0,783 & 0,474 & 0,474\\
saiga7b\_rulm\_bpe & 0,149 & 0,429 & 0,596 & 0,344 & \textbf{0,647} & 0,478 & 0,636 & 0,757 & 0,397 & 0,493\\
saiga7b\_rulm\_unigram & 0,194 & \textbf{0,432} & 0,568 & 0,313 & 0,591 & \textbf{0,587} & 0,630 & \textbf{0,789} & \textbf{0,477} & \textbf{0,509}\\
saiga7b\_rulm\_unigram\_hm & \textbf{0,198} & 0,413 & 0,584 & \textbf{0,349} & 0,533 & \textbf{0,587} & 0,578 & \textbf{0,789} & 0,475 & 0,501\\

\hline
\end{tabular}

%}
%\end{adjustbox}
%}
\end{table*}

\subsection{Human evaluation}
Since both unigram models showed similar results in automatic evaluation, we performed additional human evaluation in side-by-side comparison setting for Saiga instruction-tuned versions. We prepared 78 questions for knowledge and logic probing and generated answers using the same sampling settings. To confirm the correlation between automatic metrics and real-world task performance we included the generation results of the original Saiga 7b in model comparison set. The models were compared pair-wise in independent sessions and each question was evaluated by 15 random annotators that were selected after preliminary testing. The annotators were instructed to select the answer variants with highest factual consistency (or logical accuracy) and personal relevance (i.e. the answer with most helpful information). For the cases of equal preference the annotators had "Tie" option.

\begin{figure}
    \centering
    \includegraphics[width=0.5\textwidth]{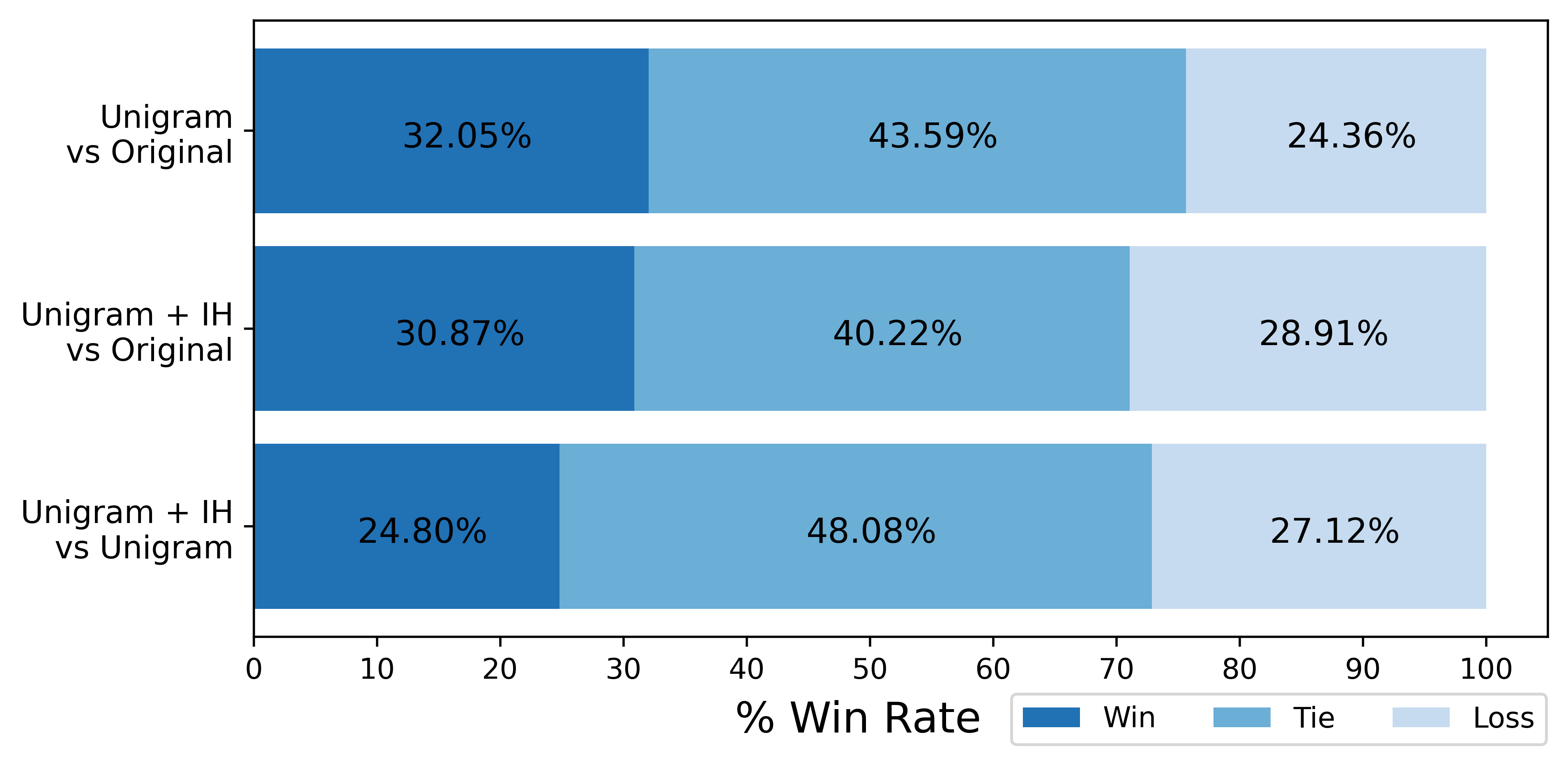}
    \caption{Human evaluation results of Saiga models}
    \label{fig:sbs}
\end{figure}

The results of human evaluation are presented in Figure \ref{fig:sbs}. As can be seen from the results, both unigram model variants outperform the original model by a significant margin. However, the annotators often found the difference in generation quality marginal, labeling the question pairs as "Tie" situation. Interestingly, exclusive LM head intialization (denoted as "+ IH") has a negative effect on instruction-tuning efficiency and in side-by-side comparison this Unigram variant loses to the simpler embedding-tuning approach, which corroborates zero-shot Russian Super Glue results from section 5.B.

\subsection{Performance}
The main benefit of tokenization optimization is guaranteed speed up of model inference and training. We evaluated the Russian text generation resource efficiency for the original LLaMa 7b and our Unigram modification. To measure the pure sampling efficiency, the generation process was guided using ForceTokensLogitsProcessor which limited the token choice at each generation timestep to predefined token sequence. The sequence was broken down by sentence count boundaries at 1,5,10 and 15 sentences. The inference was done on RTX 4090 in max performance setting with native FP16 model weigths and Flash attention 2 enabled.

\begin{figure}
    \centering
    \includegraphics[width=0.43\textwidth]{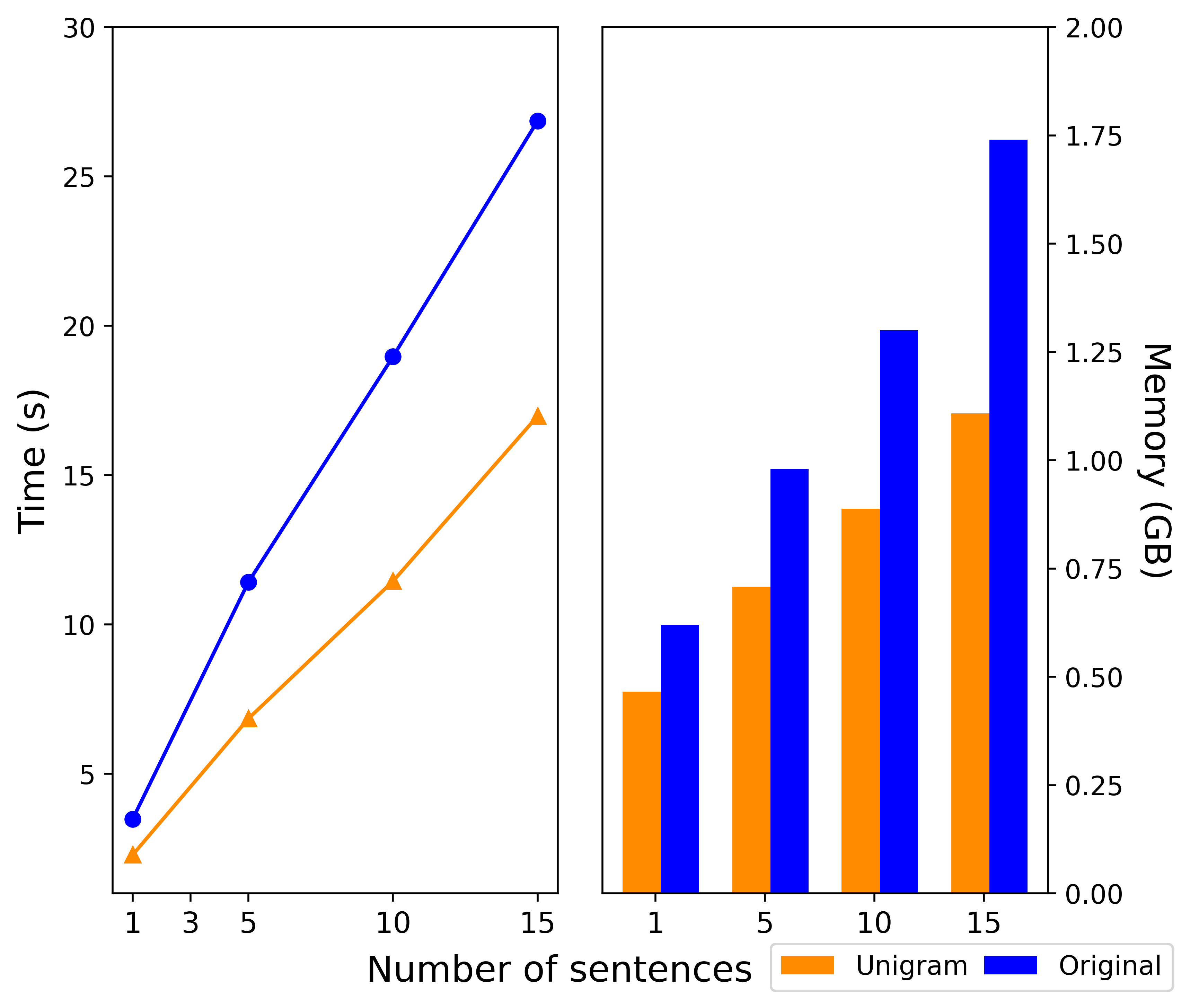}
    \caption{Resource consumption for different text lengths}
    \label{fig:pefr}
\end{figure}

Figure \ref{fig:pefr} shows the generation time and additional memory consumed for text length breakpoints. As can be seen the improvement in average token count substantially boosts the resource efficiency. The difference in time and memory consumption for short sequences (1 sentence) is marginal, however it becomes more evident with increasing text length. To produce a story of 15 sentences the generation process for our Unigram model would take approximately 17 seconds, while the original backbone would spend 27 seconds, thus achieving a $\sim$60\% speed up. Similarly, fine-tuning the model with the new tokenization on the instruct dataset is 35\% faster, taking 20 hours instead of the original 27 hours at the same training configuration. 

\section{Conclusions}
We explored the possibility of adapting large language models (LLaMA) to Russian language with vocabulary substitution method. Our tokenization experiments with Unigram and BPE algorithms showed that Unigram algorithm is better suited for Russian language, having better morphological accuracy and comparable token count. This result was also confirmed on Russian Super Glue benchmark which demonstrated considerable quality improvements for all variants of vocabulary substitution with Unigram tokenizer bringing the highest improvements. Additional Human Evaluation of Saiga instruction-tuned models indicated better instruction utilization after Unigram vocabulary substitution which has higher user preference over the original Saiga7b model. We also showed that vocabulary substitution is an efficient method for resource optimization, achieving significant speed up both for fine-tuning (35\%) and inference (up to 60\%) with memory consumption reduction. 

\section*{Acknowledgements}
The work of Mikhail Tikhomirov (general concept, method implementation, experiments) was supported by Non-commercial Foundation for Support of Science and Education "INTELLECT". The work of Daniil Chernyshev (dataset collection, result interpretation, survey) was supported by Non-commercial Foundation for Support of Science and Education "INTELLECT". The research is carried out using the equipment of the shared research facilities of HPC computing resources at Lomonosov Moscow State University.

%\vspace{12pt}
%\color{red}
%IEEE conference templates contain guidance text for composing and formatting conference papers. Please ensure that all template text is removed from your conference paper prior to submission to the conference. Failure to remove the template text from your paper may result in your paper not being published.

\end{document}